\documentclass{article}
\usepackage{paper}

\usepackage{times}
\usepackage{soul}
\usepackage{url}
\usepackage[hidelinks]{hyperref}
\usepackage[utf8]{inputenc}
\usepackage[small]{caption}
\usepackage{graphicx}
\usepackage{amsmath}
\usepackage{amsthm}
\usepackage{booktabs}
\usepackage{algorithm}
\usepackage{algorithmic}
\usepackage[switch]{lineno}

\usepackage{subfloat}
\usepackage{subcaption,booktabs}
\usepackage{bbm}
\usepackage{amssymb}

\title{Learning 3D Photography Videos via Self-supervised Diffusion on Single Images
}

\author{
Xiaodong Wang$^{1}$\thanks{Equal contribution.}
\and
Chenfei Wu$^{2}$\footnotemark[1]\and
Shengming Yin$^{2}$\and
Minheng Ni$^{2}$ \and
Jianfeng Wang$^{3}$ \and
Linjie Li$^{3}$ \and
Zhengyuan Yang$^{3}$ \and
Fan Yang$^{2}$ \and
Lijuan Wang$^{3}$ \and
Zicheng Liu$^{3}$ \and
Yuejian Fang$^{1}$ \and
Nan Duan$^{2}$\thanks{Corresponding author.}
\affiliations
$^1$Peking University\ 
$^2$Microsoft Research Asia\ 
$^3$Microsoft Azure AI\\
\emails
\{wangxiaodong21s@stu, fangyj@ss\}.pku.edu.cn,
\{chewu, v-sheyin, t-mni, jianfw, Lindsey.Li, zhengyang, fanyang, lijuanw, zliu, nanduan\}@microsoft.com,
}

\begin{document}

\maketitle
\begin{figure*}
    \centering
    \includegraphics[width=1.0\linewidth]{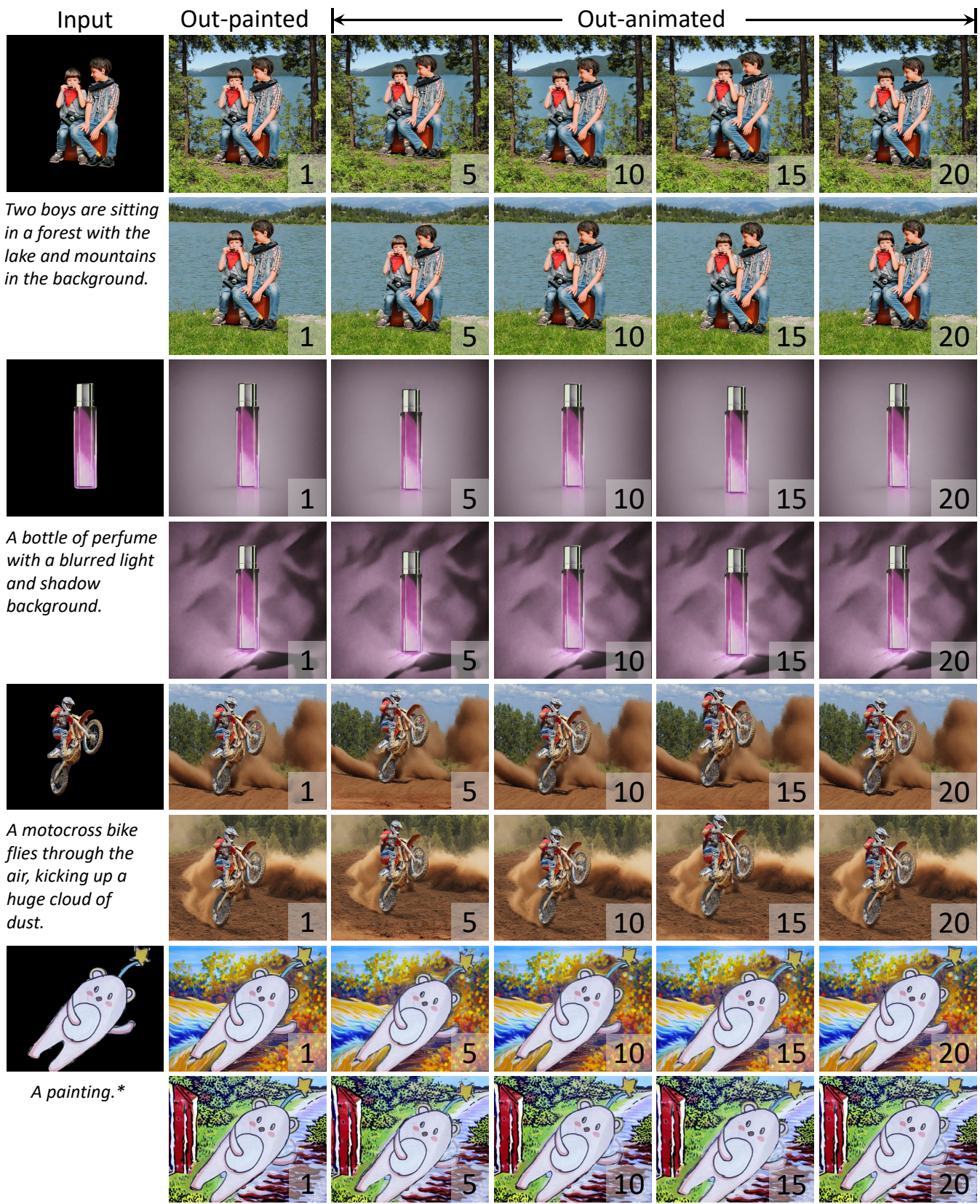}
    \caption{Illustration of our proposed out-animation task. We first outpaint the input into complete and suitable scenes based on its content and prompts, and then generate subsequent frames with 3D effects to form the out-animated videos. Our method can handle a wide variety of scenes from the open-domain, such as people, objects, advertised goods, paintings, etc. ($^*$This input comes from an artwork of a young artist Geng Jiahao, in 2022 ANOBO ``A Drop of Water with One World" exhibition. The number indicates the frame number in a 3D video.)}
    \label{fig:demo}
\end{figure*}
\begin{figure*}
    \centering
    \includegraphics[width=1.\linewidth]{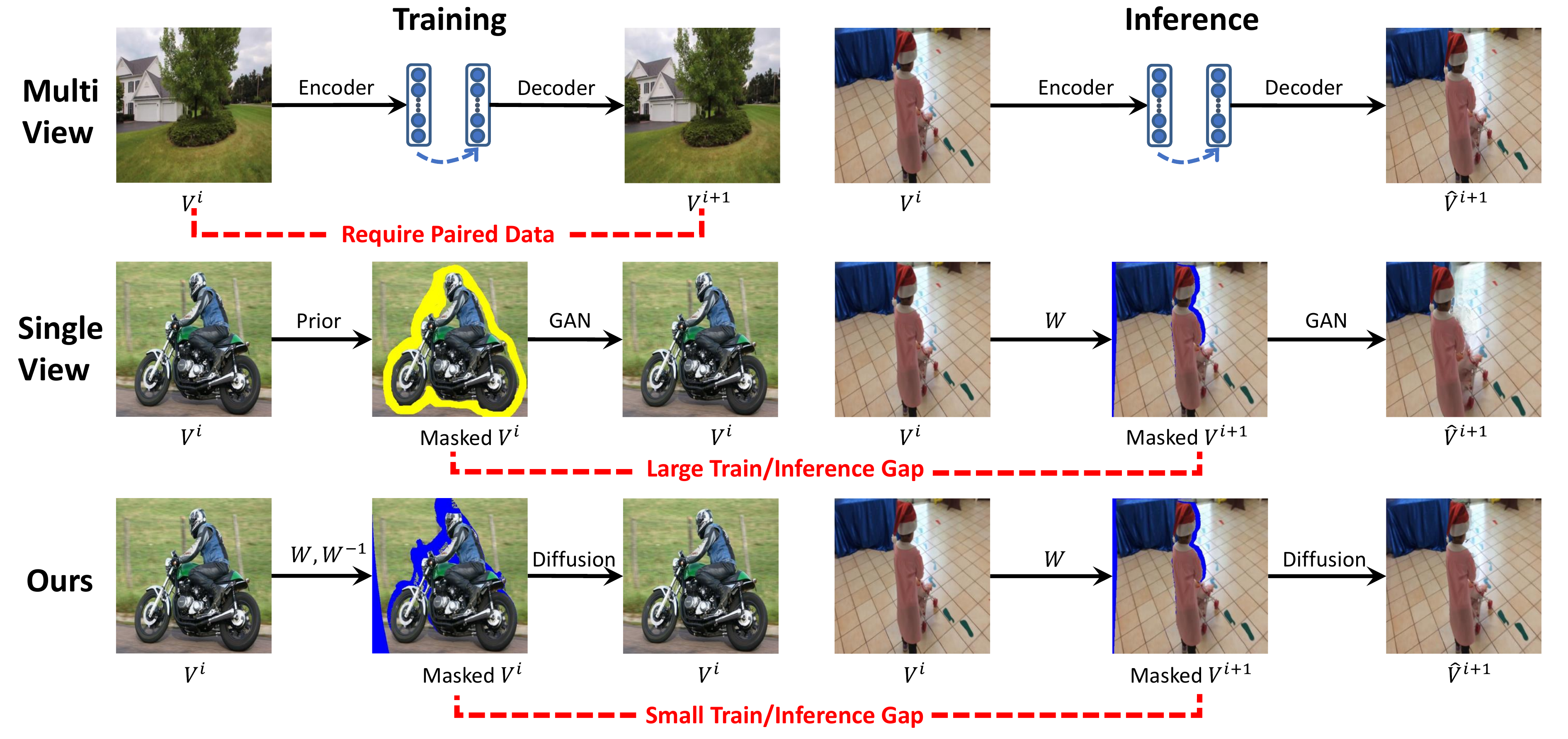}
    \vspace{-2mm}
    \caption{Comparison of different methods. Our proposed self-supervised diffusion belongs to single-view methods but is different from existing works. We incorporate a pair of 3D rendering function $W, W^{-1}$ to reduce the train/inference gap of existing single-view methods.}
    \label{fig:motiv}
\end{figure*}


\begin{abstract}
 \vspace{-2mm}

3D photography renders a static image into a video with appealing 3D visual effects. Existing approaches typically first conduct monocular depth estimation, then render the input frame to subsequent frames with various viewpoints, and finally use an inpainting model to fill those missing/occluded regions. The inpainting model plays a crucial role in rendering quality, but it is normally trained on out-of-domain data. To reduce the training and inference gap, we propose a novel self-supervised diffusion model as the inpainting module. Given a single input image, we automatically construct a training pair of the masked occluded image and the ground-truth image with random cycle-rendering. The constructed training samples are closely aligned to the testing instances, without the need of data annotation. To make full use of the masked images, we design a Masked Enhanced Block (MEB), which can be easily plugged into the UNet and enhance the semantic conditions. Towards real-world animation, we present a novel task: out-animation, which extends the space and time of input objects. Extensive experiments on real datasets show that our method achieves competitive results with existing SOTA methods.
\end{abstract}
 \vspace{-4mm}
\section{Introduction}
Recent advances in generation have delivered impressive photorealistic visual contents, such as images~\cite{rombachHighResolutionImageSynthesis2022,wuNUWAInfinityAutoregressiveAutoregressive2022,rameshHierarchicalTextconditionalImage2022} and videos~\cite{wuGODIVAGeneratingOpenDomaIn2021,villegasPhenakiVariableLength2022,singerMakeAVideoTexttoVideoGeneration2022}. 
3D photography is a special case of video generation that converts a static image into a 3D video, with applications in virtual reality and animation, attracting wide attention~\cite{shih3dPhotographyUsing2020,wilesSynSinEndtoendView2020,tuckerSingleviewViewSynthesis2020,jampaniSLIDESingleImage2021}.  It requires the model to generate a sequence of \emph{consistent} and \emph{reasonable} subsequent frames (novel views) from a starting image only by interactively changing the camera viewpoint, thus putting forward higher requirements to training data and model design.

Existing 3D photography methods can be summarized into two categories: \emph{multi-view methods} and \emph{single-view methods}, as shown in the first two rows in Fig.~\ref{fig:motiv}, respectively. Multi-view methods ~\cite{huWorldsheetWrappingWorld2021,laiVideoAutoencoderSelfsupervised2021,rockwellPixelsynthGenerating3dconsistent2021,liMineContinuousDepth2021,liSynthesizingLightField2020} usually take frames from video, multi-view images, or fake multi-view images~\cite{hanSingleViewViewSynthesis2022} as input and produce 3D representations of scenes, such as point clouds~\cite{wilesSynSinEndtoendView2020} and MPIs (multiplane images)~\cite{tuckerSingleviewViewSynthesis2020}, and then train an encoder-decoder model projecting the source view $V^i$ to target view $V^{i+1}$. As shown in the first row in Fig.~\ref{fig:motiv}, such methods need image pairs in training, but there is a lack of large-scale multi-view datasets.

To alleviate the dependency on multi-view datasets, single-view methods~\cite{jampaniSLIDESingleImage2021,shih3dPhotographyUsing2020} try to train with only RGB image data, first estimate the monocular depth~\cite{ranftlRobustMonocularDepth2022}, and then use MPIs or LDIs (layered depth images) representations to better inpaint the occluded regions of target frames. 
Early method~\cite{shih3dPhotographyUsing2020} searches the occluded regions by hard layering the discontinuities of estimated depth and inpaints the depth and RGB images separately. Further, ~\cite{jampaniSLIDESingleImage2021} extends a soft depth layering to decompose an image into the foregrounds and backgrounds, and it prefers inpainting the backgrounds for preserving foreground details. These single-view methods convert the 3D photography into an inpainting task by predefining the masked regions on image data, and these predefined masks by priors are more suitable for generation than random masks, therefore, they can generate good 3D videos only trained on single images.

However, in 3D photography for complex scenes, the predefined masks in single-view methods are not the real occluded regions caused by 3D rendering. As shown in the second row in Fig.~\ref{fig:motiv}, single-view methods first predefine the masks according to edge or depth priors (the yellow masked regions surrounding the foregrounds as in~\cite{jampaniSLIDESingleImage2021}), and then train GAN-based models to inpaint these regions towards original images. During inference, they first render the source frame $V^i$ into a masked $V^{i+1}$ at the target viewpoint via the 3D renderer $W$, as the blue masked image shows in the second row in Fig.~\ref{fig:motiv}, and then use the trained model to inpaint those occluded regions to obtain the novel view $\hat{V}^{i+1}$. Since the rendered masks are different from predefined masks, there is a large gap between training and inference, leading to some visual distortions.

To this end, we propose a novel self-supervised diffusion model, which only trains on single images but can generate high-quality 3D videos shown in Fig.~\ref{fig:motiv}. To align inpainting masks towards real occluded masks, we use the same 3D renderer $W$ to compose the cycle-rendering $(W, W^{-1})$. In cycle-rendering, we first use $W$ to randomly render the image $V^i$ to a virtual image at a nearby viewpoint, and then we use $W^{-1}$ to render the virtual image back to the viewpoint of $V^i$. As a result, we obtain the masked $V^i$ with occluded regions. Further, we utilize a conditional diffusion model that denoises cycle-rendered masked images into source images. We regard the masked images as the conditions, and the source images as the ground-truth images of our diffusion model, respectively. During inference, we can inpaint the masked $V^{i+1}$ via our trained diffusion model to generate the novel view $\hat{V}^{i+1}$. This self-supervised way effectively reduces the gap between training and inference, guaranteeing the high quality of generated 3D videos.
To fully leverage the semantic information of masked images, we present a Masked Enhanced Block (MEB) to better embed them into the denoising UNet. Specifically, the masked images and occluded masks are fed into the MEBs via two stacked spatially adaptive normalization layers~\cite{parkSemanticImageSynthesis2019}. 
By leveraging the enhanced module to inpaint masked images, our model outputs higher fidelity results with fewer visual distortions. 
To further prompt the visual quality and diversity of novel views, we can transfer the text-to-image knowledge into our self-supervised diffusion model, relieving the pressure of the inpainting process.

Towards the real application of animation, we further present a novel task: out-animation, which requires the model to generate a video that extends the space and time of input objects (or selected parts of an image). We propose a two-stage pipeline: We first perform the image-outpainting with the same denoising network to generate consistent and reasonable scenes for input objects according to their contents and text prompts, and then sample the 3D videos via the trained self-supervised diffusion model. Experiments on novel view synthesis and image-outpainting of real datasets validate the effectiveness of our method.  

We summarize our contributions as follows:
 \begin{itemize}
    \item We propose a novel self-supervised diffusion model, which trains only on single images but can generate high-quality 3D photography videos.
    \item We propose MEB, a Masked Enhanced Block that leverages the unmasked image conditions for the denoising process of our diffusion model.
    \item We present a novel task: out-animation, and adapt the proposed diffusion models to the new task. Experiments on real datasets validate the effectiveness of our models.
 \end{itemize}
 \vspace{-4mm}

\section{Related Work}
\subsection{3D Photography} 
\paragraph{Multi-view methods.} Many methods~\cite{hanSingleViewViewSynthesis2022,huWorldsheetWrappingWorld2021,laiVideoAutoencoderSelfsupervised2021,rockwellPixelsynthGenerating3dconsistent2021,liMineContinuousDepth2021,liSynthesizingLightField2020,wang3DMomentsNearDuplicate2022a,mildenhallNeRFRepresentingScenes2020} learn to predict the 3D representations (such as NeRf~\cite{mildenhallNeRFRepresentingScenes2020}, point clouds~\cite{wilesSynSinEndtoendView2020}, and MPIs~\cite{tuckerSingleviewViewSynthesis2020} using multi-view supervision, so they assume many views such as related two views or multi-views can be used to train the models.
\cite{tuckerSingleviewViewSynthesis2020} first applies the MPI representations from single image input and renders them to novel views, and then predicts the target views via multi-view supervision. 
SynSin~\cite{wilesSynSinEndtoendView2020} proposes a novel point cloud render that transfers the latent 3D point cloud features into the target views, and rendered features are passed a refinement network to generate target predictions. To alleviate the lack of large-scale in-the-wild multi-view datasets, AdaMPI~\cite{hanSingleViewViewSynthesis2022} trains an inpainting network via warp-back strategy to construct fake multi-view datasets and then trains an MPI-based model.

\paragraph{Single-view methods.}
Single-view methods~\cite{tuckerSingleviewViewSynthesis2020,shih3dPhotographyUsing2020,jampaniSLIDESingleImage2021,kopfOneShot3D2020,niklaus3DKenBurns2019} only require single view images with LDIs (layered depth images)~\cite{shih3dPhotographyUsing2020} representations. Most single-view methods estimate the dense monocular depths and fill in the predefined occluded regions. 3d photo~\cite{shih3dPhotographyUsing2020} makes full use of the three inpainting modules that separately inpaint the edges, depth, and color images to predict the impressive novel views. SLIDE~\cite{jampaniSLIDESingleImage2021} proposes a soft layering to separate an image into foregrounds and backgrounds, and then it prefers inpainting the color and depth of backgrounds, therefore, it better preserves the foreground details.

\subsection{Image Synthesis}
\paragraph{Text-conditional synthesis.} Text-to-image recently becomes popular due to the remarkable success of autoregressive models and diffusion models. Parti~\cite{yuScalingAutoregressiveModels2022} and NUWA-Infinity~\cite{wuNUWAInfinityAutoregressiveAutoregressive2022} are two-stage approaches, which first compress an image into discretized latent space, and autoregressively predict discrete image tokens based on text tokens. Diffusion-based methods generate the images via a denoising network conditioned on text representations, such as DALL$\cdot$E2~\cite{rameshHierarchicalTextconditionalImage2022}, Imagen~\cite{sahariaPhotorealisticTexttoImageDiffusion2022}, and Stable-Diffusion~\cite{rombachHighResolutionImageSynthesis2022}.

\paragraph{Semantic image synthesis.} This task aims to create images from semantic segmentations, and the challenge is to generate better images in terms of visual fidelity and spatial alignment.
SPADE~\cite{parkSemanticImageSynthesis2019} is the most popular method to achieve promising images by easily introducing a spatially adaptive normalization layer. From this conditional normalization perspective, Methods~\cite{liuLearningPredictLayouttoimage2019,tanEfficientSemanticImage2021,wangImageSynthesisSemantic2021,zhuSeanImageSynthesis2020} designed more tailored ways to embed the semantic masks.

Recently, researchers also pay more attention to \emph{image-inpainting} and \emph{image-outpainting}. Inpainting aims to fill missing regions in images, while outpainting tries to extend the images. Some text-conditional image synthesis methods could directly adapt the outpainting or inpainting. Stable-Diffusion and DALL$\cdot$E2 can be applied on both outpainting and inpainting tasks. 
The core of 3D photography is similar to image-inpainting, but inpaints the occluded regions for moved 3D target views. The MEB in our UNet is inspired by the various embedding ways of semantic masks, but we utilize the diffusion models rather than GAN-based models.
\section{Methodology}
In this paper, we propose a novel self-supervised diffusion model for learning 3D photography. We are the first to leverage the diffusion model for 3d photography, and the quality of novel views will benefit from the randomness continuously involved by noise at each step. We organize the rest of this section as follows: We first review the conditional diffusion models. Then, we present our self-supervised diffusion model. After that, we will present the out-animation task, and introduce our pipeline for achieving this task.
\begin{figure*}
    \centering
    \includegraphics[width=\linewidth]{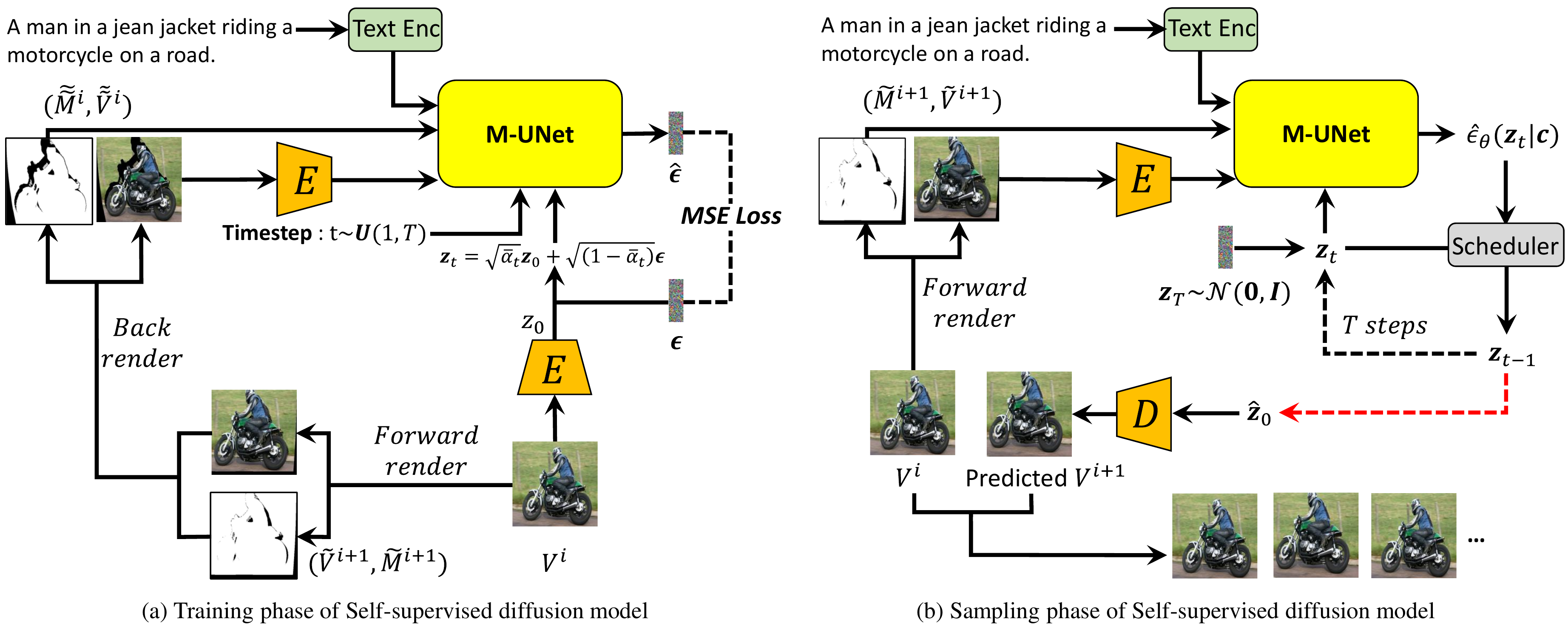}
    \caption{The overall framework. (a) A self-supervised way to train the diffusion model to inpaint the rendered masked regions which align closely to real occlusions, and (b) inferences a consistent and realistic 3D video by our effective model with the small train-inference gap.}
    \label{fig:main}
\end{figure*}

\subsection{Preliminaries}
We briefly review the theory of conditional diffusion models~\cite{hoDenoisingDiffusionProbabilistic2020}. Conditional diffusion models are latent variable models that aim to learn the form of $p_\theta(\mathbf{z}_0|\mathbf{c})$ while the conditional data follows $q(\mathbf{z}_0|\mathbf{c})$, where $c$ is the condition. The joint distribution $p_\theta(\mathbf{z}_{0:T}|\mathbf{c})$ is called the \emph{reverse process}, which is a Markov chain with learned Gaussian transitions starting at $p(\mathbf{z}_T)=\mathcal{N}(\mathbf{z}_T;\mathbf{0},\mathbf{I})$, as shown below:
\begin{align} 
    p_\theta(\mathbf{z}_{0:T}|\mathbf{c}) := p(\mathbf{z}_T) \prod \limits_{t=1}^T p_\theta(\mathbf{z}_{t-1}|\mathbf{z}_t,\mathbf{c}), \\
    p_\theta(\mathbf{z}_{t-1}|\mathbf{z}_t,\mathbf{c}):=\mathcal{N}(\mathbf{z}_{t-1};\mathbf{\mu}_\theta(\mathbf{z}_t,\mathbf{c},t),\mathbf{\Sigma}_\theta(\mathbf{z}_t,\mathbf{c},t))
\end{align}
where $t$ is an arbitrary timestep.
The approximate posteriors $q(\mathbf{z}_{1:T}|\mathbf{z}_0)$, called the \emph{forward process}, is fixed as a Markov chain that gradually adds Gaussian noise to the data obeying a variance scheduler $\beta_1,\dots,\beta_T$:
\begin{align}
    q(\mathbf{z}_{1:T}|\mathbf{z}_0):=\prod \limits_{t=1}^T q(\mathbf{z}_{t}|\mathbf{z}_{t-1}), \\
    q(\mathbf{z}_{t}|\mathbf{z}_{t-1}):= \mathcal{N}(\mathbf{z}_t; \sqrt{1-\beta_t}\mathbf{z}_{t-1},\beta_t \mathbf{I})
\end{align}
We define $\alpha_t :=1-\beta_t$ and $\overline{\alpha}_t:= \prod_{s=1}^t\alpha_s$. Then we sample $\mathbf{z}_{t}$ at the timestep $t$ in closed form as follows:
\begin{equation}
    q(\mathbf{z}_{t}|\mathbf{z}_0) = \mathcal{N}(\mathbf{z}_{t};\sqrt{\overline{\alpha}_t}\mathbf{z}_0,(1-\overline{\alpha}_t)\mathbf{I})
\end{equation}
The conditional diffusion models are trained to minimize the variational lower bound (VLB), and the objective is equivalent to a denoising process as follows:
\begin{equation}
    \mathcal{L}_{t-1}=\mathbbm{E}_{\mathbf{z}_0,\mathbf{c},\mathbf{\epsilon} 
 \sim \mathcal{N}(\mathbf{0},\mathbf{I}),t} \Big[\left\| \epsilon-\epsilon_{\theta}(\mathbf{z}_t,t,\mathbf{c})\right\|_2^2 \Big]
\end{equation}
where $\mathcal{L}_{t-1}$ is the loss function at the timestep $t-1$, and $\epsilon_\theta$ is the denoising network. 

\subsection{Self-supervised Diffusion Model}
To address the lack of large-scale multi-view data and generate high-quality 3D videos, we present the self-supervised diffusion model. The two crucial parts of our model are the self-supervised strategy for denoising and the M-UNet (a modified UNet with several masked enhanced blocks).

\paragraph{Self-supervised strategy.} Previous works~\cite{liInfiniteNatureZeroLearningPerpetual2022,hanSingleViewViewSynthesis2022} attempted to construct the training pairs from images, i.e. occluded images and ground-truth images, to get rid of the requirement for multi-view datasets. But they only trained the traditional inpating models to inpaint the RGBD images, which are prone to predict inconsistency results in occluded regions. To this end, we take the best of the strong synthesis capability of the diffusion model to facilitate the high visual quality of these occluded regions. 

Inspired by~\cite{liInfiniteNatureZeroLearningPerpetual2022}, we use a cycle-rendering way to achieve the self-supervised diffusion process in our model.
The overall self-supervised denoising process is shown in Fig.~\ref{fig:main} (a). The cycle-rendering includes a forward rendering and a back rendering. Given an input image, we regard it as the $i$-th frame $V^i$. In forward rendering, We first estimate the depth image $D^i$ by monocular depth estimation (such as~\cite{ranftlRobustMonocularDepth2022}). Then, we randomly sample a nearby viewpoint, which is a virtual viewpoint with a relative camera pose $T=[R|t]$, where $R$ is the rotation matrix and $t$ is the translation matrix. We render the frame $V^i$ to a pseudo next frame $\Tilde{V}^{i+1}$ at that virtual viewpoint as follows:
\begin{equation}
    (\Tilde{V}^{i+1}, \Tilde{D}^{i+1}) = \mathcal{W}(V^i, D^i, T)
\end{equation}
where $\mathcal{W}$ is the same 3D render process as ~\cite{liuInfiniteNaturePerpetual2021a}. In forward rendering, we can find some occluded regions deriving from the depth image $\Tilde{D}^{i+1}$, marked by $\Tilde{M}^{i+1}$, so we obtain the pair $(\Tilde{V}^{i+1}, \Tilde{M}^{i+1})$ as shown in Fig.~\ref{fig:main} (a). However, we do not have the ground truth of this frame. So, we render this occluded next frame back to the original viewpoint via a back rendering as follows:
\begin{equation}
    (\Tilde{\Tilde{V}}^{i}, \Tilde{\Tilde{D}}^{i}) = \mathcal{W}\big((\Tilde{V}^{i+1} \cdot \Tilde{M}^{i+1},\Tilde{D}^{i+1}\cdot \Tilde{M}^{i+1}),T^{-1}\big)
\end{equation}
where $T^{-1}$ is the inverse camera pose, and mask $\Tilde{M}^{i+1}$ is element-wise multiplied with the RGBD frame $(\Tilde{V}^{i+1},\Tilde{D}^{i+1})$, and then the render way is same as the forward rendering. After that, we can obtain a masked image with mask $(\Tilde{\Tilde{V}}^{i},\Tilde{\Tilde{M}}^{i})$ at the same viewpoint with original frame $V^i$. These pairs $\{(\Tilde{\Tilde{V}}^{i},\Tilde{\Tilde{M}}^{i}), V^i\}$ enable the diffusion model to inpaint the occluded regions only on single images by a self-supervised way. We regard the masked images $\Tilde{\Tilde{V}}^{i}$ as image conditions, and the image $V^i$ is the ground truth of our diffusion model. We will encode the masked images to latent features, and input them into M-UNet with mask and a text prompt as conditions. We encode the ground truth to latent features $\mathbf{z}_0$, and we add the noise to latent features as follows:
\begin{equation}
    \mathbf{z}_t = \sqrt{\overline{\alpha}_t}\mathbf{z}_0 + \sqrt{(1-\overline{\alpha}_t)}\mathbf{\epsilon}
\end{equation}
The noisy latent features $\mathbf{z}_t$ are fed into M-UNet with the timestep $t$ to predict the added noise $\hat{\epsilon}$. At every timestep, the denoising objective is the MSE loss between $\hat{\epsilon}$ and $\epsilon$.

\begin{figure}
    \centering
    \includegraphics[width=1.0\linewidth]{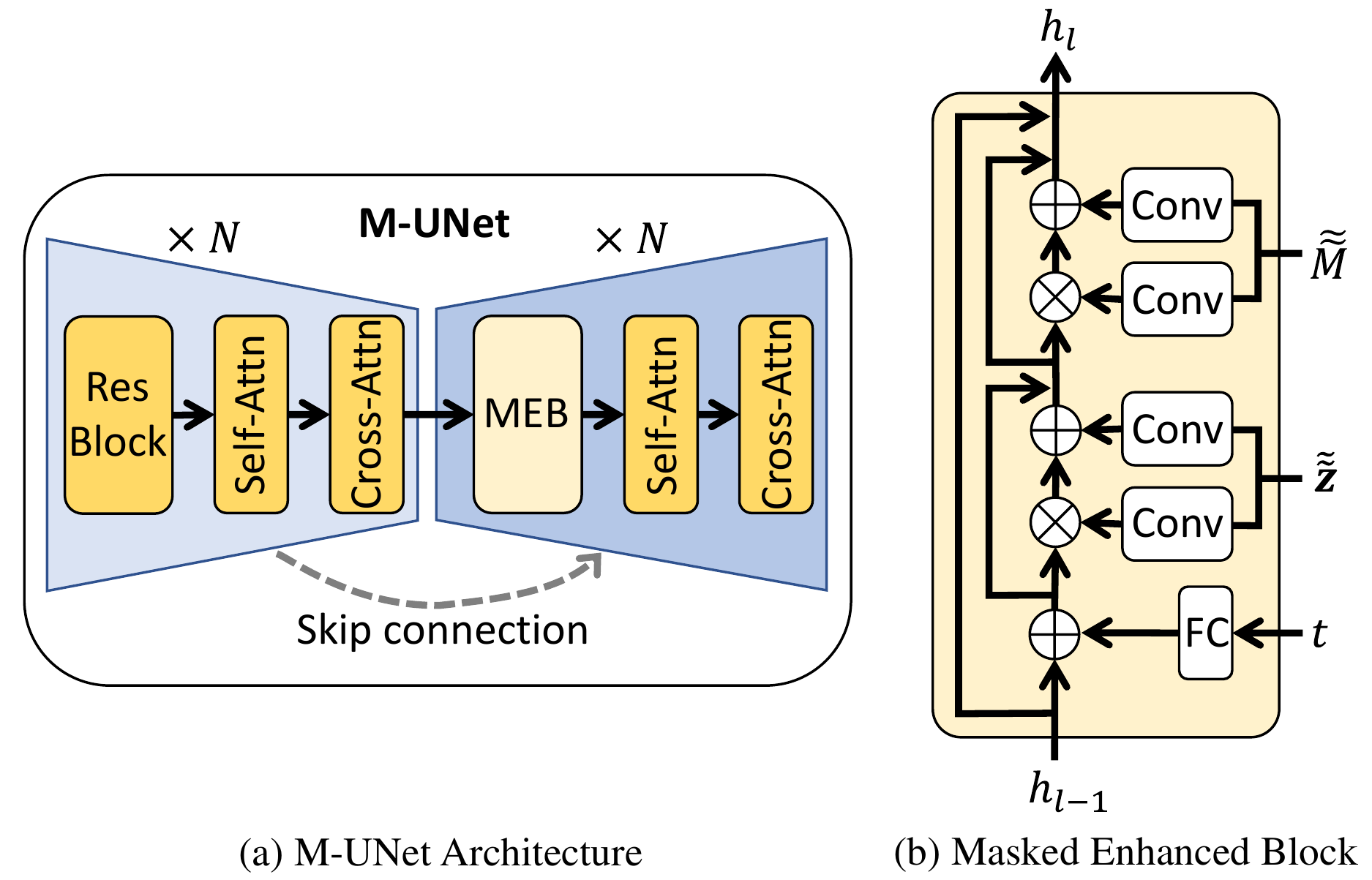}
    \caption{Illustrations of M-UNet (a) and Masked Enhanced Block (MEB) (b). MEB is a key component of M-UNet.}
    \label{fig:unet}
\end{figure}

\paragraph{M-UNet.} Based on UNet, we propose M-UNet with a set of Masked Enhanced Blocks (MEBs). Traditional inpainting networks struggle to predict consistency contents for occluded regions, and one reason is that unmasked spatial information suffers loss in image-to-image networks, as well as the concatenation in diffusion models~\cite{wangSemanticImageSynthesis2022}. The diffusion model is very flexible for conditions, so we design a masked enhanced block to fully leverage the unmasked regions. As shown in Fig.~\ref{fig:unet} (a), we take the same architecture of UNet as in Stable-diffusion~\cite{rombachHighResolutionImageSynthesis2022}, but we design several MEBs for the downblocks of the UNet, since image conditions would suffer less loss when embedded into downblocks~\cite{wangSemanticImageSynthesis2022}. We utilize two stacked spatially normalization layers to embed the masked image features $\Tilde{\Tilde{\mathbf{z}}}^i$ and the mask $\Tilde{\Tilde{M}}^i$ for enhancing the spatial information, and add the timestep $t$ embedding as normal, as shown in Fig.~\ref{fig:unet} (b). 
We formulate the stacked spatially normalization layers as follows:
\begin{equation}
    \mathbf{f}^i = \gamma_{M}(\Tilde{\Tilde{M}}) 
 \cdot \big( \gamma_\mathbf{z}(\Tilde{\Tilde{\mathbf{z}}})\cdot\texttt{Norm}(\mathbf{f}^{i-1})+\beta_\mathbf{z}(\Tilde{\Tilde{\mathbf{z}}}) \big) + \beta_\mathbf{M}(\Tilde{\Tilde{M}})
\end{equation}
where the $\mathbf{f}^{i-1}$ and $\mathbf{f}^i$ are the input and output features. $\gamma(\cdot)$ and $\beta(\cdot)$ are the convolution layers to map latent features or masks into high-level spatially-adaptive features and add them with input features. The enhancement denotes that the embedded mask further enhances the spatial information of unmasked regions. 
We add the skip-connection at the start of MEB, the start of the first embed layer, and the start of second embed layer, as shown in Fig.~\ref{fig:unet} (b). We insert the MEB into each downblock to fully leverage the image conditions. The stacked normalization layers are crucial parts of M-UNet, which show better performance than other embedding ways.
\paragraph{Training.} The training process is shown in Fig.~\ref{fig:main} (a). We denote the M-UNet as $\epsilon_\theta$, the noise added from Gaussian distribution as $\epsilon$, the mask as $\Tilde{\Tilde{M}}^{i}$, and the occluded image features as $\Tilde{\Tilde{\mathbf{z}}}^{i}=E(\Tilde{\Tilde{V}}^{i})$ encoded by KL-VAE encoder. The text prompt features are denoted as $\mathbf{p}$ encoded by CLIP text encoder~\cite{radfordLearningTransferableVisual2021a}. We regard conditions of M-UNet as the set of $\mathbf{c}=\{\Tilde{\Tilde{M}}^{i},\Tilde{\Tilde{\mathbf{z}}}^{i},\mathbf{p}\}$, and then we learn the conditional self-supervised diffusion model via:
\begin{equation}
    \mathcal{L}(V^i)=\mathbbm{E}_{E(V^i),\mathbf{c},\mathbf{\epsilon} 
 \sim \mathcal{N}(\mathbf{0},\mathbf{I}),t} \Big[\left\| \mathbf{\epsilon}-\epsilon_{\theta}(\mathbf{z}_t,t,\mathbf{c})\right\|_2^2 \Big]
\end{equation}
with $t$ uniformly sampled from $\{1,\dots,T \}$, $\epsilon_{\theta}$ is optimized during training. To support the classifier-free guidance sampling, we randomly drop 10\% text prompts during training.

\paragraph{Sampling.} We use the classifier-free guidance for sampling videos with our diffusion model as shown in Fig.~\ref{fig:main} (b). Given the $i^{th}$ frame $V^i$, we first forward render it to obtain an occluded image and mask $(\Tilde{V}^{i+1},\Tilde{M}^{i+1})$. The occluded image is encoded to latent features $\Tilde{\mathbf{z}}^{i+1}$. We use the blank text $\emptyset$ with the mask and occluded features as the extra conditions, and normal conditions $\mathbf{c}=\{\Tilde{M}^{i+1},\Tilde{\mathbf{z}}^{i+1},\mathbf{p} \}$ consists of the text prompt, the mask, and occluded image features. The guidance way in the sampling procedure can be formulated:
\begin{equation}
    \hat{\epsilon}_\theta(\mathbf{z}_t,\mathbf{c}) = \epsilon_\theta(\mathbf{z}_t,\mathbf{c})+s\cdot\big(\epsilon_\theta(\mathbf{z}_t,\mathbf{c})-\epsilon_\theta(\mathbf{z}_t,\Tilde{M}^i,\Tilde{\mathbf{z}}^{i+1},\emptyset)\big)
\end{equation}
where $s$ is the guidance scale. After predict the noise at $t$ timestep, we can obtain the latent features $\mathbf{z}_{t-1}$ as follows:
\begin{equation}
    \mathbf{z}_{t-1} = \frac{1}{\sqrt{\alpha_t}}\big(\mathbf{z}_t - \frac{1-\alpha_t}{\sqrt{1-\overline{\alpha}_t}} \hat{\epsilon}_\theta(\mathbf{z}_t|\mathbf{c})\big) + \mathbf{\Sigma}_\theta(\mathbf{z}_t|\mathbf{c})^{\frac{1}{2}}\mathbf{n}
\end{equation}
where $\mathbf{n} \sim \mathcal{N}(\mathbf{0},\mathbf{I})$.
At the start of sampling, $\mathbf{z}_T$ is the pure Gaussian noise, and we denoise each $\mathbf{z}_t$ to $\mathbf{z}_{t-1}$ following the linear time scheduler $T,T-1,\dots,0$. Finally, we obtain the predicted features $\mathbf{\hat{z}}_0$ as the red dotted arrow shows in Fig.~\ref{fig:main} (b), and $\mathbf{z}_0$ can be decoded to a predicted $\hat{V}^{i+1}$ frame with a single pass through KL-VAE decoder $D$. Then, we can randomly sample a 3D video autoregressively.

\begin{figure}
    \centering
    \includegraphics[width=\linewidth]{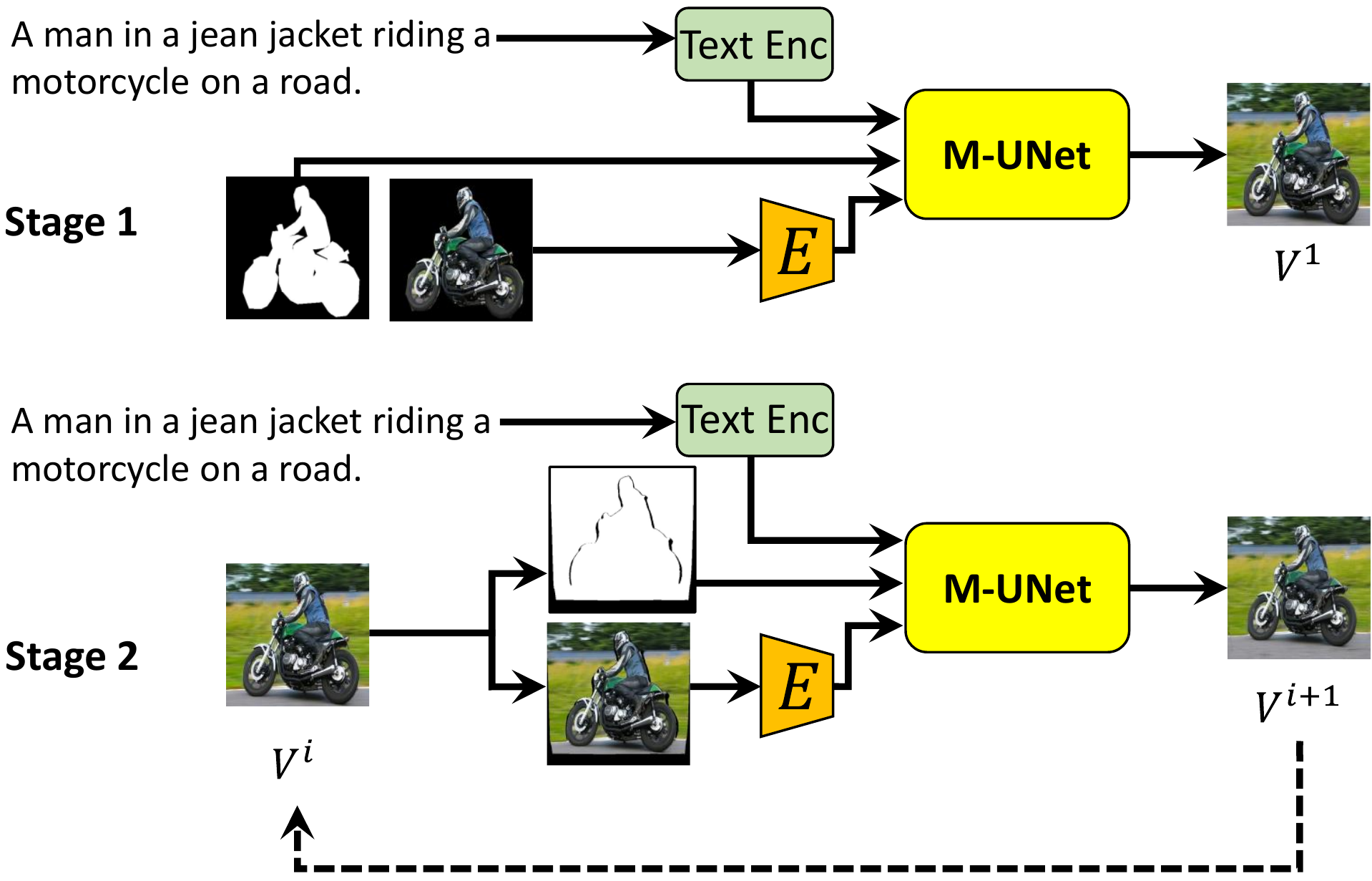}
    \caption{Two-stage pipeline for out-animation.}
    \label{fig:outanim}
\end{figure}

\subsection{Out-animation}
3D photography can generate the subsequent frames following starting static input to compose a 3D video, but most contents of these frames are the same as the input. In real applications, generating more high-quality content provides more convenience for creative work of the real world, such as text-to-image~\cite{rombachHighResolutionImageSynthesis2022}, image-outpainting~\cite{wuNUWAInfinityAutoregressiveAutoregressive2022,changMuseTextToImageGeneration2023}. So, an efficient 3D video generation is supposed to have the ability to extend the input image along both space and time dimensions.
\paragraph{Definition.} Given input objects (or selected parts of an image), out-animation aims to extend input objects along both space and time dimensions and outputs an animated video.
\paragraph{Our pipeline.} We decompose the out-animation task into two sub-tasks: image-outpainting and 3D photography. Our pipeline consists of two stages as shown in Fig.~\ref{fig:outanim}, and both only depend on the denoising M-UNet. 


The stage-1 pipeline is shown at the top of Fig.~\ref{fig:outanim}. The input objects may have arbitrary shapes. To generate desired complete images, we can train a diffusion model based on M-UNet to denoise input objects to the ground truth images. The training data can be collected from semantic segmentation datasets (such as MSCOCO~\cite{caesarCocostuffThingStuff2018}). Similar to our strategy for inpainting 3D photography, we regard source images as the ground truth and segmented objects and masks as image conditions. In addition, text prompts are conditions for this diffusion model. We use MEBs to embed the segmented objects and masks into M-UNet. For inference of a new scene, we input the masks, objects, and text prompts to the trained diffusion model. The objects at pixel-level will be encoded by KL-VAE encoder and be embedded to M-UNet with masks. Then, after $T$ steps denoising process, we can obtain the final latent features, and decode them to the first frame $V^1$ via KL-VAE decoder. The stage-2 pipeline regards synthesis results came from stage-1 as the starting frame, and iteratively renders the current frame $V^i$ and then generates the next frame $V^{i+1}$ as shown in the bottom of Fig.~\ref{fig:outanim}.

Our method is different from the 3d photography methods~\cite{shih3dPhotographyUsing2020,jampaniSLIDESingleImage2021,liInfiniteNatureZeroLearningPerpetual2022} based on depth-inpainting, since we only utilize depth images in our rendering processes. We aim to leverage the diffusion model to predict the novel views, rather than training a complex model or extra model for depth-inpainting. 

\section{Experiments}
For a fair comparison with state-of-the-art methods, we evaluate the synthesis results on two datasets, RealEstate10k (RE10K)~\cite{zhouStereoMagnificationLearning2018a} which provides about 10 million frames derived from about 80k video clips of static scenes, and MannequainChalenge (MC)~\cite{liLearningDepthsMoving2019} with more than 170k frames derived from about 2k YouTube videos. To validate the effectiveness of our M-UNet, we also evaluate the image-outpainting in MSCOCO~\cite{caesarCocostuffThingStuff2018}.

\paragraph{Baselines and Metrics.} 
For \textbf{novel view synthesis}, we quantitatively and qualitatively compared our method with recent state-of-the-art methods for which code is released: SynSin~\cite{wilesSynSinEndtoendView2020}, Single-image MPI (SMPI)~\cite{tuckerSingleviewViewSynthesis2020}, 3d-photo~\cite{shih3dPhotographyUsing2020}, and AdaMPI~\cite{hanSingleViewViewSynthesis2022}. In our experiments, we evaluate the released pretrained models. SynSin and SMPI models were trained on RealEstate10K. 3d-photo, AdaMPI, and ours were trained on MSCOCO. For a fair comparison, 3d-photo, AdaMPI, and our method all use MiDaS~\cite{ranftlRobustMonocularDepth2022} for depth estimation. SynSin and SMPI models are trained on RE10K, which randomly samples source and target frames from 57K training clips. Our model loaded the weights from Stable-diffusion, and only trained the 3D photography on MSCOCO dataset. We use the same intrinsic matrices, source camera poses and target camera poses for all methods. Following 3d-photo~\cite{shih3dPhotographyUsing2020}, we measure the accuracy of the predicted target views with ground-truth images using three metrics including LPIPS, PSNR, SSIM.
For \textbf{image-outpainting}, we quantitatively and qualitatively compared with recent method: Stable-Diffusion~\cite{rombachHighResolutionImageSynthesis2022}, the pre-trained model SDM-v1.4 and the finetuned version on MSCOCO by our implement. We measure the FID, IS, and CLIP similarity of outpainted images with ground-truth images.

\begin{figure*}
    \centering
    \includegraphics[width=\linewidth]{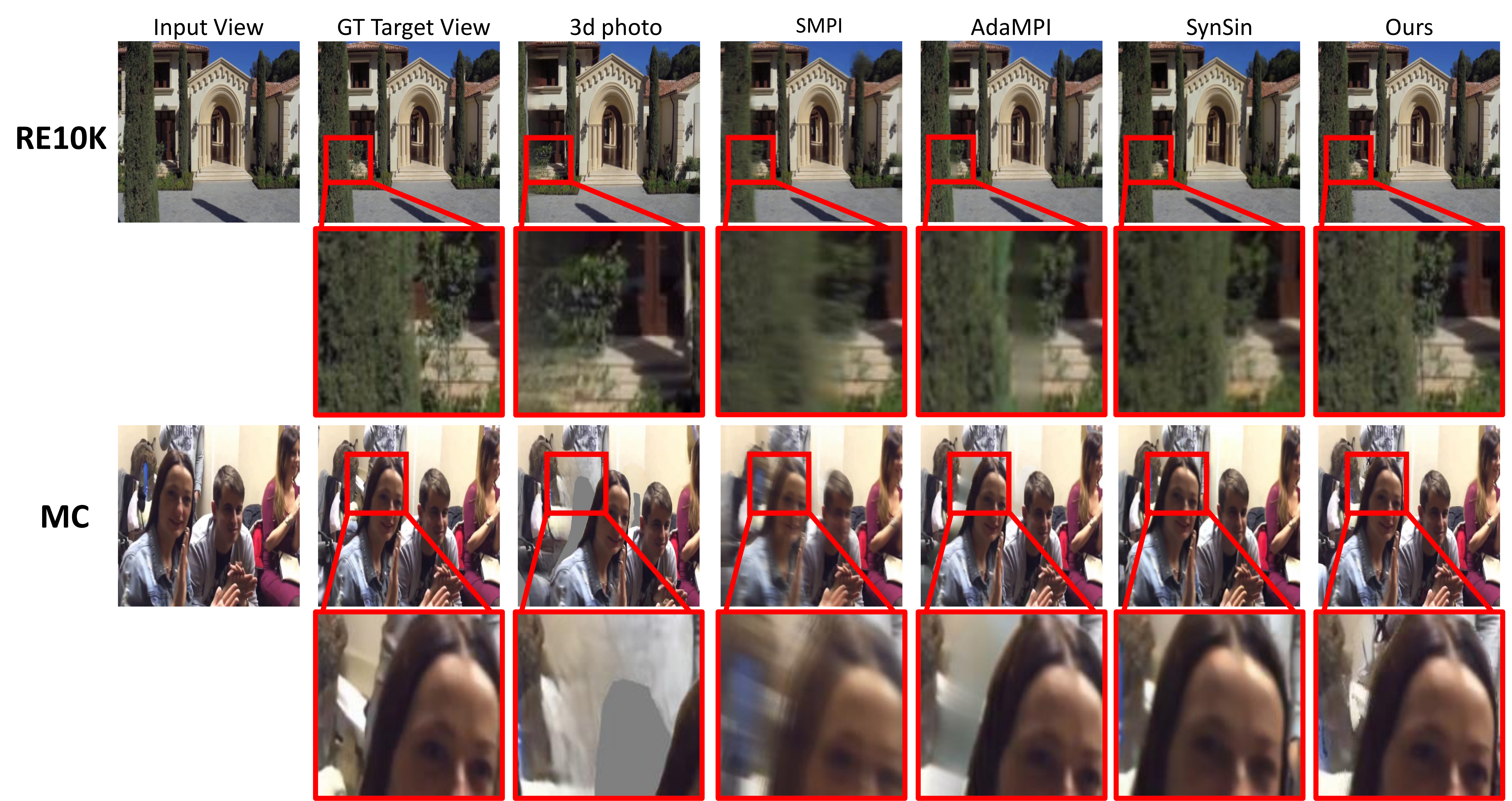}
    \caption{Sample visual results on benchmarks. Our method generates clear visual content and preserves details better than others.}
    \label{fig:vis1}
\end{figure*}

\paragraph{Quantitative Results on RealEstate10k.} RealEstate10k (RE10K)~\cite{zhouStereoMagnificationLearning2018a} is a video dataset consisting of 10K YouTube videos of static scenes. We randomly sample 1K video clips from test set for evaluation. We use the given camera intrinsics and extrinsics following RE10K. Specifically, we choose the first frame ($t$=1) from each test clip as the source view and consider the fifth ($t$=5) frame and tenth ($t$=10) frames as target views. We compute the evaluation metrics between predictions and ground-truth views. The results are shown in Tab.~\ref{tab:real}. We report our results of the trained model loaded from Stable-diffusion and the model only trained on MSCOCO from scratch. 3d-photo and our method do not need video datasets and perform zero-shot on RE10K dataset because both are only trained on MSCOCO. Compare with multi-view methods (SynSin and SMPI), our method outperforms SynSin on almost all results and shows better LPIPS than SMPI, indicating that we can achieve high realistic results than others. However, since we will generate new better content for occluded regions, the structural similarity with GT target views would slightly drop, but show better visual effects. Compared zero-shot performance with 3d-photo and AdaMPI, we perform better results than 3d-photo by a large margin and competitive results with AdaMPI.

\begin{table}
    \centering
    \resizebox{\linewidth}{!}{
    \begin{tabular}{lcccccccc}
        \toprule
          & & & \multicolumn{2}{c}{LPIPS$\downarrow$} & \multicolumn{2}{c} {PSNR$\uparrow$ } & \multicolumn{2}{c}{SSIM$\uparrow$} \\
        \cmidrule(lr){4-5} \cmidrule(lr){6-7} \cmidrule(lr){8-9}
        Method & Video &Zero-shot & $t=5$ & $t=10$ & $t=5$ & $t=10$ & $t=5$ & $t=10$ \\
        \midrule
        3d-photo & & \checkmark & \underline{0.209} & \underline{0.266} & \underline{16.00} & \underline{15.27} & \underline{0.43} & \underline{0.41} \\
        SynSin & \checkmark  &   & 0.063 & 0.097 & 24.42 & 21.73 & 0.81 & 0.71  \\
        SMPI &\checkmark & & 0.055 & 0.106 & \textbf{26.90} & \textbf{23.32} & \textbf{0.87} & \textbf{0.78} \\
        AdaMPI & & \checkmark &0.056 & 0.100  & 25.83  &22.01  & 0.84 & 0.73  \\
        \midrule
        Ours (Scratch) &  & \checkmark  &\underline{0.052}  &\underline{0.099}  &\underline{25.11}  &\underline{21.18}  & \underline{0.83} & \underline{0.72}  \\
        Ours &  & \checkmark  & \underline{\textbf{0.049}} & \underline{\textbf{0.095}} & \underline{25.35} & \underline{21.36} & \underline{0.84} & \underline{0.72}  \\
        
        \bottomrule
    \end{tabular}
    }
    \caption{Results on RealEstate10K. $\downarrow$ denotes higher is better, and $\downarrow$ lower is better, underline masks the single-view methods. We evaluate the target views at two timesteps $t$=5 and $t$=10, size=256$\times$256.}
    \label{tab:real}
\end{table}

\begin{table}
    \centering
    \resizebox{\linewidth}{!}{
    \begin{tabular}{lcccccccc}
        \toprule
          & & & \multicolumn{2}{c}{LPIPS$\downarrow$} & \multicolumn{2}{c} {PSNR$\uparrow$ } & \multicolumn{2}{c}{SSIM$\uparrow$} \\
        \cmidrule(lr){4-5} \cmidrule(lr){6-7} \cmidrule(lr){8-9}
        Method &Video &Zero-shot & $t=3$ & $t=5$ & $t=3$ & $t=5$ & $t=3$ & $t=5$ \\
        \midrule
        3d-photo & & \checkmark & \underline{0.495} & \underline{0.590} & \underline{11.88} & \underline{11.23} & \underline{0.29} & \underline{0.27} \\
        SynSin & \checkmark & \checkmark   & 0.304 & 0.404 & 15.53 & 13.83 & 0.44 & 0.37  \\
        SMPI &\checkmark &\checkmark & 0.349 & 0.453 & \textbf{16.26} & \textbf{14.56} & 0.47 & \textbf{0.41} \\
        AdaMPI & &\checkmark & 0.298 & 0.393 & 16.21 & 14.35 & \textbf{0.49} & \textbf{0.41} \\
        \midrule
        Ours (Scratch) &  & \checkmark  & \underline{0.277}  &\underline{0.372}  &\underline{15.87}  &\underline{13.87}  &\underline{0.47}  & \underline{0.37}  \\
        Ours &  & \checkmark  & \underline{\textbf{0.272}} & \underline{\textbf{0.367}} & \underline{16.01} & \underline{13.99} & \underline{0.47} & \underline{0.38} \\
        
        \bottomrule
    \end{tabular}
    }
    \caption{Results on MannequinChallenge, underline masks the single-view methods. Two timesteps $t$=3 and $t$=5, size=256$\times$256.}
    \label{tab:mc}
\end{table}

\paragraph{Quantitative Results on MannequinChallenge (MC).} MC is a video dataset of video clips of people freezing in diverse and natural poses, which is collected and processed similarly to RE10K. Similar to~\cite{jampaniSLIDESingleImage2021}, we randomly sample 200 video clips from the test set, and we choose the  first frame ($t$=1) as the source view, and the third frame ($t$=3) and fifth frame ($t$=5) as target views. The results are shown in Tab.~\ref{tab:mc}. All methods did not finetune on MC dataset, so all reports are zero-shot results. Our method outperforms 3d-photo and SynSin on all results by a large margin. Our method achieved the best LPIPS and comparable results for PSNR and SSIM with SMPI and AdaMPI, indicating the best perceptual similarity and good structure similarity. 

\begin{table}
    \centering
    \resizebox{0.8\columnwidth}{!}{
    \begin{tabular}{lccc}
        \toprule
        Method & FID$\downarrow$ & IS$\uparrow$ & CLIP-SIM$\uparrow$\\
    \midrule
         SD & 16.77 & 35.13 & 0.3212   \\
         SD-FT & 13.01  & 34.09  & 0.3112   \\
    \midrule
         Ours (Scratch) & 16.51  & 35.95  & 0.3218  \\
         Ours w/ cross-fusion & 11.33  & 38.40  & 0.3214  \\
         Ours w/o mask & 10.66 & 37.59 & 0.3216 \\
         Ours & \textbf{10.65} & \textbf{38.61} & \textbf{0.3226} \\
         
    \bottomrule
    \end{tabular}
    }
    \caption{Outpointing results on MSCOCO.}
    \label{tab:coco}
\end{table}

\paragraph{Quantitative Results on MSCOCO.}
MSCOCO-2017~\cite{caesarCocostuffThingStuff2018} contains a total of 172 classes: 80 thing classes, 91 stuff classes, and 1 unlabeled class. We focus on thing classes and utilize pixel-level annotations for training. We regard the regions belonging to thing classes as the input objects regions, and models are supposed to outpaint the remaining regions. For each image, we utilize the pixel-level annotations to construct the binary mask to indicate the known objects and unknown regions, and captions are regarded as the text prompts. We filter the original dataset to an outpainting dataset containing 117266 training images and 4952 test images. We evaluate the pretrained Stable-diffusion model (SD), a finetuned model on MSCOCO (SD-FT), and a variant of ours trained from scratch. Since Stable-diffusion does not support the training for outpainting, we only finetune it for the text-to-image task. Evaluation results are shown in Tab.~\ref{tab:coco}. The finetuned Stable-diffusion (SD-FT) achieves a better FID than pretrained model (SD), but degrades on IS and CLIP-SIM. It is reasonable that the generated images from the finetuned model are closer to the original images, while the quality, diversity, and text relevance are slightly worse. Our model significantly reduces the FID score to 10.65, and achieves better IS and CLIP-SIM than Stable-diffusion models. Without the pretrained weights, our model can still achieve better outpainting results than SD on all metrics. We also compare with the cross-fusion in MEB, introduced two learnable weights~\cite{zhuSeanImageSynthesis2020} to embed masks, and a variant of our method without masks. Our embedding way shows better quality, diversity, and text relevance, and performs slightly worse if it does not utilize masks.

\paragraph{Qualitative Results.}

\begin{figure}
    \centering
    \includegraphics[width=\linewidth]{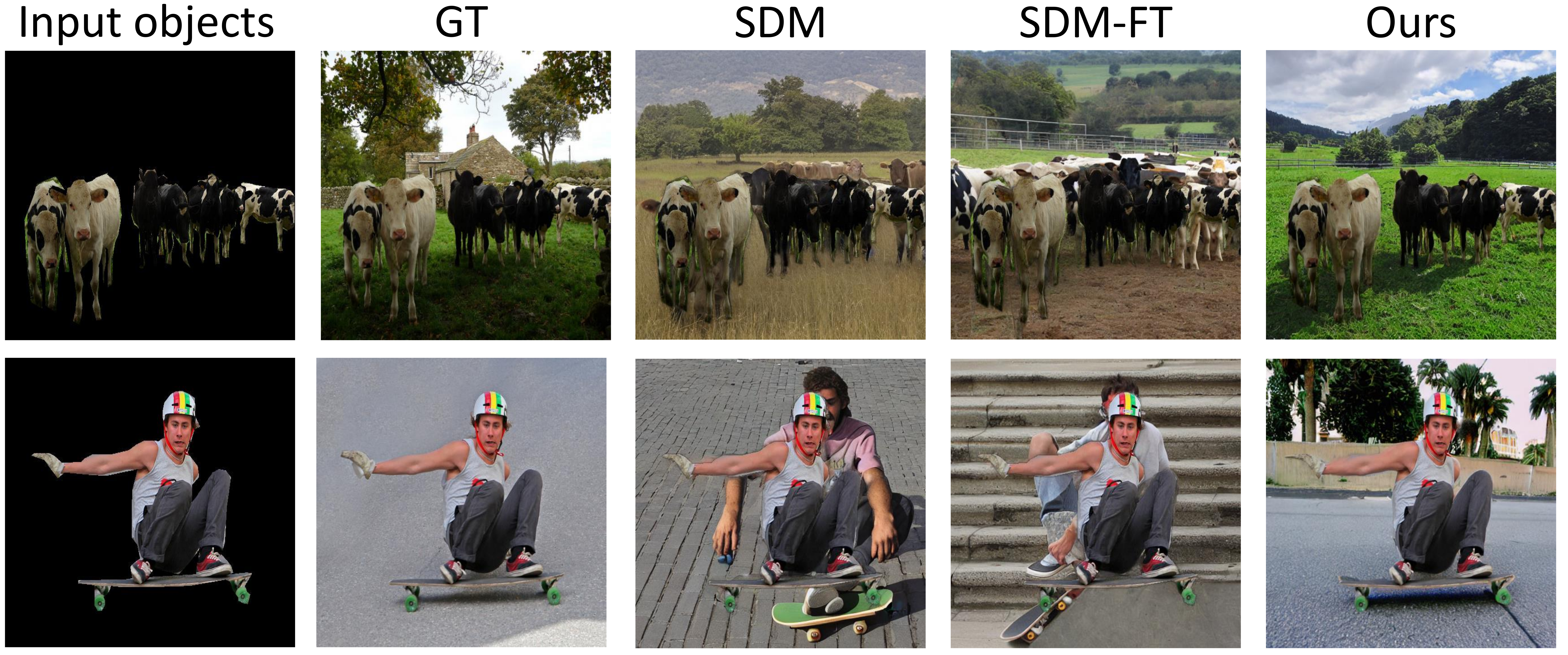}
    \caption{Outpainted results on COCO, size=512$\times$512. Our method achieves better consistent and realistic results than Stable-diffusion.}
    \label{fig:vis2}
    \vspace{-3mm}
\end{figure}

We qualitatively compare the novel view synthesis in RE10K and MC, and outpainting in COCO. As shown in the first two rows in Fig.~\ref{fig:vis1}, the goal is to generate the better target view given the input view when the camera moves. The first two rows show sample results on RE10K, and the second row shows the details in the same regions that need to be generated. Such as the right occlusion regions of the tree, 3d-photo produces results with large visual distortions and other methods generate blurrier results, while our method generates the best realistic results and preserves the structures better. We show another sample in a more challenging benchmark MC in the last two rows. 3d-photo generates worse results with more artifacts and the wrong position, and other methods also generate blurrier results in occlusion regions, while our method generates the new clear contents in these regions and preserves the clear details for other regions. We compare outpainted sample results in COCO as shown in Fig.~\ref{fig:vis2}. We use the same captions to generate the scenes for all methods. Compared with SDM and SDM-FT, our method generates more consistent and high-fidelity results.

\section{Conclusion}
To reduce the train-inference gap in 3D photography training on single images, a novel self-supervised diffusion model is proposed that can generate high-quality 3D videos from single images. We first generate the masked regions that are closely aligned to real occluded regions in 3D rendering and then train a diffusion model with masked enhanced modules to inpaint these regions. Towards the real application of animation, we present the out-animation, which extends the space and time of input objects. Experimental results on real datasets validate the effectiveness of our method.

\section*{Ethical Statement}
3d photography and out-animation generate new content and carry risks related to deceptive and otherwise harmful content. As technology improves, it will mistake generated images for authentic ones. And more research needs be done to change the societal biases in training data.

\bibliographystyle{named}
\bibliography{paper}

\end{document}